\documentclass[letterpaper, conference]{IEEEtran}
\IEEEoverridecommandlockouts
\usepackage[utf8]{inputenc}
\usepackage[T1]{fontenc}
\usepackage{graphicx}
\usepackage{amsmath,amssymb}
\usepackage{booktabs}
\usepackage{siunitx}
\usepackage{hyperref}
\usepackage{algorithm,algorithmic}
\usepackage{booktabs,siunitx,threeparttable}
\title{\vspace{0.15in}LASER: Level-based Asynchronous Scheduling and Execution Regime for Spatiotemporally Constrained Multi-Robot Timber Manufacturing}
\author{Zhenxiang Huang$^{*}$, Lior Skoury, Tim Stark, Aaron Wagner,\\
Hans Jakob Wagner, Thomas Wortmann, and Achim Menges%
\thanks{This research has been supported by the Deutsche Forschungsgemeinschaft (DFG) under Germany’s Excellence Strategy EXC 2120/1 – 390831618.} 

\thanks{Authors are with the Institute for Computational Design and Construction (ICD) and Cluster of Excellence IntCDC at University of Stuttgart, Germany. }

\thanks{$^{*}${\tt\small zhenxiang.huang@icd.uni-stuttgart.de}}%
}

\makeatletter
\def\ps@IEEEtitlepagestyle{%
  \def\@oddhead{\parbox{\textwidth}{\centering \scriptsize Accepted for publication in the 2026 IEEE International Conference on Robotics and Automation (ICRA)}\hfill}%
  \def\@oddfoot{\parbox{\textwidth}{\centering \linespread{1.0}\selectfont\scriptsize \copyright~2026 IEEE. Personal use of this material is permitted. Permission from IEEE must be obtained for all other uses, in any current or future media, including reprinting/republishing this material for advertising or promotional purposes, creating new collective works, for resale or redistribution to servers or lists, or reuse of any copyrighted component of this work in other works.}\hfill}%
}
\makeatother
\begin{document}

\maketitle
\begin{abstract}
Automating large-scale manufacturing in domains like timber construction requires multi-robot systems to manage tightly coupled spatiotemporal constraints, such as collision avoidance and process-driven deadlines. This paper introduces \textbf{LASER} (\textbf{L}evel-based \textbf{A}synchronous \textbf{S}cheduling and \textbf{E}xecution \textbf{R}egime), a complete framework for scheduling and executing complex assembly tasks, demonstrated on a screw-press gluing application for timber slab manufacturing. Our central contribution is to integrate a barrier-based mechanism into a constraint programming (CP) scheduling formulation that partitions tasks into spatiotemporally disjoint sets, which we define as ``levels.'' This structure enables robots to execute tasks in parallel and asynchronously within a level, synchronizing only at level barriers, which guarantees collision-free operation by construction and provides robustness to timing uncertainties. To solve this formulation for large problems, we propose two specialized algorithms: an iterative temporal-relaxation approach for heterogeneous task sequences and a bi-level decomposition for homogeneous tasks that balances workload. We validate the LASER framework by fabricating a full-scale 2.4\,m $\times$ 6\,m timber slab (Fig. \ref{fig:teaser}) with a two-robot system mounted on parallel linear tracks, successfully coordinating 108 subroutines and 352 screws under tight adhesive time windows. Computational studies show our method scales steadily with size compared to a monolithic approach.
\end{abstract}


\section{Introduction}
\label{sec:introduction}

The increasing demand for automation in manufacturing and construction has positioned multi-robot systems (MRS) as a powerful solution for efficiency, scalability, and safety. In large assemblies, many steps can run in parallel if spatial interference and process windows are respected. Realizing this potential, however, requires addressing tightly coupled challenges in multi-robot task allocation and scheduling, motion planning in shared workspaces, and execution under uncertainty. These challenges are magnified in industrial settings, where complex spatiotemporal constraints, bespoke parts at scale, and strict quality criteria demand schedules that are both \emph{collision-aware} and \emph{deadline-aware} by construction.

\begin{figure}[h!]
\centering
\includegraphics[width=0.98\linewidth]{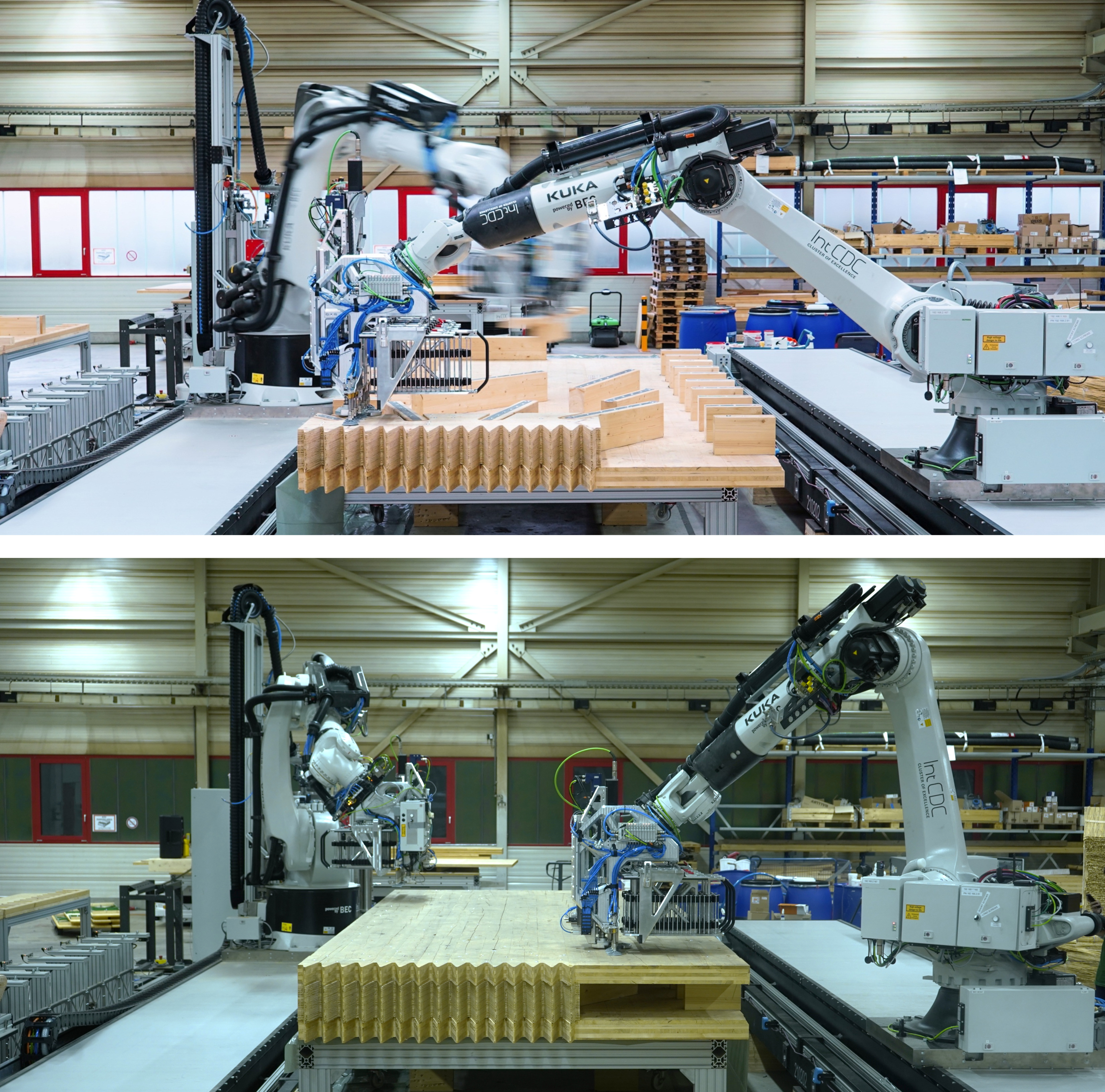}

\caption{Timber slab prototype manufacturing using a two-robot system with our proposed framework.}
\label{fig:teaser}
\end{figure}

Timber fabrication serves as a low-carbon alternative to steel and concrete, and its sustainability potential can be amplified through scalable, automated robotic manufacturing. We focus on \emph{screw–press gluing}, a connection technique that applies localized bonding pressure via screw tapping, eliminating the need for massive and costly press machines. As a traditional crafting method, this process is labor-intensive due to its dense screwing pattern (one screw per \(220\,\mathrm{cm}^2\) in our system); naively automating with a single robot is often infeasible at scale because the number of screws can violate adhesive time windows. In a hollow timber slab system, two cross-laminated timber (CLT) plates sandwich discrete timber beams for material-efficient load transfer. To manufacture such a slab segment, efficient multi-robot collaboration must respect: (i) spatial collision-avoidance of robots and payloads, and (ii) adhesive \emph{open} and \emph{close} windows that couple glue application, placement, and all subsequent screws. They are the primary drivers of schedule and execution design.

This paper presents LASER (Level-based Asynchronous Scheduling and Execution Regime), a complete framework (Fig. 2) to systematically address task scheduling and execution for spatiotemporally constrained multi-robot manufacturing, demonstrated here on the screw-press gluing process for timber slab assemblies. The main innovation of our scheduling formulation is to integrate \emph{levels} as time–space partitions of tasks whose robot workspace occupancies do not conflict. Robots execute asynchronously within a level and synchronize only at \emph{level barriers}. This lightweight mechanism yields collision-free operation by schedule design, robust to execution delays and timing uncertainty.

\begin{figure}[h!]
\centering
\includegraphics[width=1.0\linewidth]{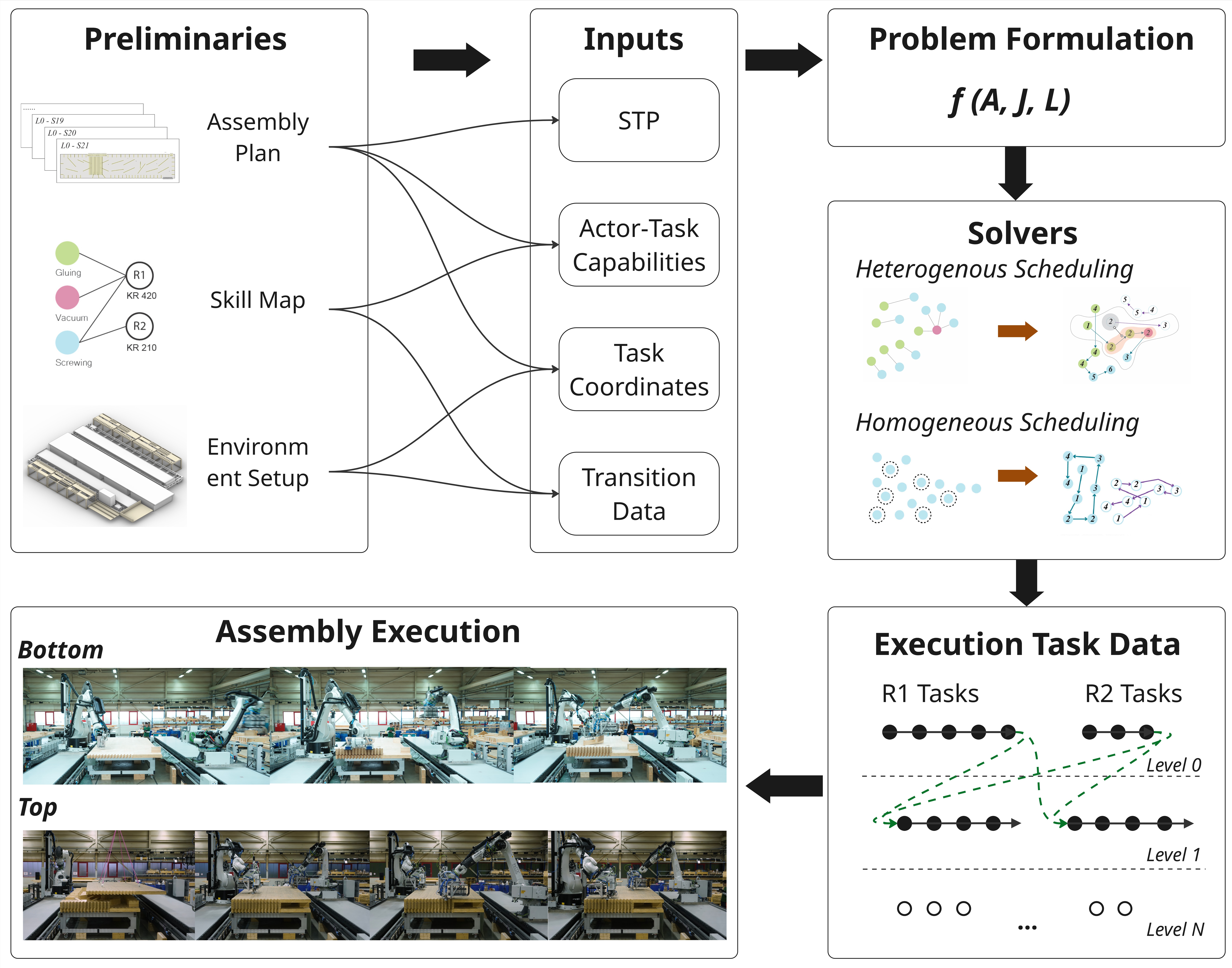}

\caption{The overall workflow of our work. We process preliminary data as inputs to our scheduling formulation, which is solved by two specific algorithms for heterogeneous and homogeneous problems. We convert the scheduled results into level-associated task objects in a semi-centralized control framework and validate our approach in a physical slab assembly.}
\label{fig:workflow}
\end{figure}

We make the following specific contributions: \textbf{1) Constraint Programming (CP) scheduling formulation with level barriers} that embeds spatial safety and adhesive time windows as constraints while minimizing makespan and level compactness. \textbf{2) Two efficient scheduling algorithms:} an iterative temporal-relaxation algorithm with heuristic task clustering for the \emph{heterogeneous} bottom assembly session; and a bi-level decomposition that translates the problem into set partitioning and vehicle routing problems (VRP) for the \emph{homogeneous} top screw-pressing session. \textbf{3) Barrier-based, semi-centralized execution} that queues level-scoped subroutines to each robot, enabling parallel execution and sparse synchronization. \textbf{4) Physical validation and scalability study} via a full-scale \SI{2.4}{\meter}\,$\times$\,\SI{6}{\meter} prototype featuring 34 internal beams, 15 levels, 108 subroutines, and 352 screws with a 2-robot setup, alongside a computational study over 8 design variants demonstrating scalability.

%We make the following specific contributions:
%\begin{enumerate}
    %\item \textbf{Constraint Programming (CP) scheduling formulation with level barriers.} We formulate scheduling problem with a \emph{level-based barrier mechanism} that embeds spatial safety and adhesive time windows as constraints while minimizing makespan and level compactness. 
    %\item \textbf{Two efficient scheduling algorithms.} 
    %We propose an iterative temporal-relaxation algorithm with heuristic task clustering for the \emph{heterogeneous} bottom assembly session; and a bi-level decomposition algorithm that translates the main scheduling formulation into well-known set partitioning and vehicle routing problems for the \emph{homogeneous} top screw-pressing session.
    %\item \textbf{Barrier-based, semi-centralized execution.} We integrate the schedule with a semi-centralized dispatcher that queues level-scoped subroutines to each industrial robot, enabling parallel execution within level and sparse synchronization under barriers.
    %\item \textbf{Physical validation and scalability study.} We fabricate a full-scale \SI{2.4}{\meter}\,$\times$\,\SI{6}{\meter} prototype featuring \textbf{34} internal beams, \textbf{15} levels, \textbf{108} dispatched subroutines, and \textbf{352} screws with a 2-robot setup. A computational study over \textbf{8} different design variants demonstrates scalability and efficiency.
%\end{enumerate}

\section{Related Work}

Our framework sits at the intersection of three major research threads within multi-robot systems—industrial manufacturing, scheduling with spatiotemporal constraints, and asynchronous execution.

\subsection{Multi-Robot Systems for Industrial Manufacturing}
Early work established multi-robot assembly as a distinct industrial challenge with coordinated fixtures, tool sequencing, and collision-aware co-manipulation. The survey by Marvel \emph{et al.} synthesizes strategies and metrics for industrial multi-robot assembly, covering fixtureless methods, synchronization schemes, and evaluation protocols~\cite{Marvel2018}. In automotive spot-welding, Pellegrinelli \emph{et al.} developed integrated approaches that jointly consider cell design and off-line motion planning~\cite{PELLEGRINELLI2014Welding, Pellegrinelli2017}. Beyond automotive, manufacturing tasks such as aircraft assembly~\cite{Tereshchuk2021} and brick layering~\cite{uavmasonry2025} introduce heterogeneous collaboration with tool change and precedence requirements. Large multi-station industrial lines motivate hierarchical allocation and scheduling; Zhou \emph{et al.} present a stepwise MS-MRTA method validated on a production case~\cite{Zhou2022}. For broader assembly planning at the structure level, discrete optimization (ILP/IQP) has been used to co-design assembly sequences and resource usage for multi-robot teams~\cite{Culbertson2019}.

\subsection{Multi-Robot Task Scheduling with Spatiotemporal Constraints}
Gerkey and Matari\'{c} differentiate allocation types by robot/task multiplicity and time extension~\cite{Gerkey2004}; Korsah \emph{et al.} add interrelated utilities and constraints~\cite{Korsah2013}; and Nunes \emph{et al.} target precedence and time windows explicitly~\cite{Nunes2017}. Exact/heuristic schedulers that couple space and time have been proposed, TERCIO~\cite{GombolayTRO2018} handles tight spatial interference and deadlines at scale. Under execution uncertainty, hierarchical methods decouple per-robot policies from conflict resolution while preserving time-window feasibility~\cite{Choudhury2020}. These works are grounded in temporal reasoning: Simple Temporal Networks (STNs) and their controllability/dispatchability properties~\cite{Dechter1991,Morris2005}. In practical settings, constraint programming (CP) offers compact encodings of precedence, resource sharing, and spatial safety using interval variables and global constraints. The CP-SAT solver \cite{ortoolsCPSAT}, part of Google OR-Tools, supports scalable formulations for precedence, resource sharing, and spatial safety.

\subsection{Asynchronous Multi-Robot Execution Frameworks}
Bridging plans to execution, recent frameworks advocate partial-order, asynchronous dispatch to address execution uncertainty without dense time-stamped synchronization. APEX-MR post-processes task/motion plans to build a temporal plan graph that enables robust asynchronous execution for cooperative assembly and demonstrates substantial speedups on long-horizon tasks~\cite{huang2025apexmr}. The Bidirectional Temporal Plan Graph (BTPG) further introduces switchable dependencies to reduce wait time while maintaining safety guarantees during execution~\cite{SuAAAI2024}. At the middleware level, asynchronous multi-arm/multi-robot execution with continuous collision checking has been implemented in MoveIt~2, providing a practical substrate for independent yet safe trajectory execution under a central scheduler~\cite{MoveIt2Async2024}. While prior works like ~\cite{huang2025apexmr} enable asynchronous execution, our level-based barrier provides a novel, integrated mechanism that simplifies the execution framework by encoding collision avoidance directly into the schedule structure, thereby obviating the need for continuous collision checking or complex temporal plan graphs during execution.

%%%%%%%%%%%%%%%%%%%%%%%%%%%%%%%%%%%%%%%%%%%%%%%%%%%%%%%%%%%%%%%%%%%%%%%%%%%%%%%%
\section{Preliminaries}
\label{sec:prelims}

The cooperative slab assembly for our multi-robot system provides a set of preliminaries for the scheduling model. They form the basis for the problem formulation in Sec.~\ref{sec:problem}. They include the design of slab assembly plan, a library of manipulation skills and subprograms mapped to robots, and the environment setup with robot cell layout as a result of a co-design process to ensure the feasibility of fabrication.

\subsection{Assembly Plan and Task Hierarchy}
\label{sec:primitives}
The timber slab assembly plan provides a list of elements $E = \{e_1, e_2, \ldots, e_N\}$ for the bottom session tasks. Each element is associated with a set of task primitives, including gluing, picking, placing, and screw-pressing.
\begin{itemize}
\item \textbf{Element Types:} For the bottom session, elements can be either linear beams (manipulated by a screw effector with an integrated gripper) or planar crowns (handled by a vacuum gripper). Each element is associated with a list of glue lines and screws to be fastened.
\item \textbf{Task Primitives:} The tasks for the bottom session are heterogeneous, involving a mix of picking, placing, gluing, and screw-pressing. For the top session, tasks are homogeneous, consisting solely of high-density screw-pressing to secure the top CLT plate.
\end{itemize}
The assembly plan specifies an unordered set of elements, subject to a small number of precedence constraints. For instance, a crown must be placed before its adjacent beams to facilitate proper slot-in placement. Additionally, for the top session, screws are divided into a \textbf{priority set} and a \textbf{reinforcement set}, where all screws in the priority set must be fastened before those in the reinforcement set.

\subsection{Environment Setup and Cell Layout}
\label{sec:environment}
Our fabrication platform consists of two industrial robots, a KUKA KR 420 R3330 (Robot 1) and a KUKA KR 210 R3100 (Robot 2), each mounted on a 10.7-meter linear axis. The robots are positioned on opposite sides of a central movable worktable to facilitate cooperative tasks. This setup specifies the kinematics and geometries of all robots and static obstacles, as well as the placement of the workpiece in the shared workspace. The cell co-design ensures every task pose is kinematically reachable by at least one robot, a critical assumption for our formulation. 

\subsection{Manipulation Skills and Subprograms}
\label{sec:subprograms}
Each task primitive maps to a manipulation skill corresponding to a pre-defined, feedback-driven subprogram on the KUKA KRL controller. Our platforms feature a distinct division of labor based on payload capacity: Robot 1 (KUKA KR 420) is equipped with a gluing effector, a heavy-payload vacuum gripper, and a screw effector with an integrated gripper, enabling it to perform all tasks including gluing, placing large crowns, and screw-pressing. In the meantime, Robot 2 (KUKA KR 210) is equipped solely with a screw effector with an integrated gripper; it is limited to placing linear beams and screw-pressing but utilizes a more powerful motor to reduce cycle time and maximize utilization. We collect average execution times for these skills and encode them into the scheduling model to minimize the estimated makespan.

%\subsection{Manipulation Skills and Subprograms}
%\label{sec:subprograms}
%Each task primitive is mapped to a manipulation skill, which corresponds to a pre-defined subprogram on the KUKA KRL controller. Our robotic platforms have different payload capacities, leading to a distinct division of labor:
%%\begin{itemize}
%\item \textbf{Robot 1 (KUKA KR 420)}: Equipped with a gluing effector, a heavy-payload vacuum gripper, and a screw effector with an integrated gripper. This robot can perform all tasks, including gluing, picking/placing large crowns, and screw-pressing.
%\item \textbf{Robot 2 (KUKA KR 210)}: Equipped with a screw effector with an integrated gripper, this robot is limited to picking/placing linear beams and screw-pressing. To maximize its utilization, the screw effector uses a more powerful motor to reduce cycle time.
%%\end{itemize}
%During execution, each skill is performed by a feedback-driven subprogram on the KUKA controller. We collect average time data for each skill and encode these durations in the scheduling model to minimize the estimated makespan.

\section{Problem Formulation}\label{sec:problem}
In this section, we formulate the scheduling of multi-robot screw-press gluing as a \textbf{Constraint Programming (CP)} problem. Building upon the discussion in Sec.~\ref{sec:prelims}, we derive the following key inputs for our CP model, as summarized in Fig. ~\ref{fig:input}:

\begin{figure}[h!]
\centering
\includegraphics[width=1.0\linewidth]{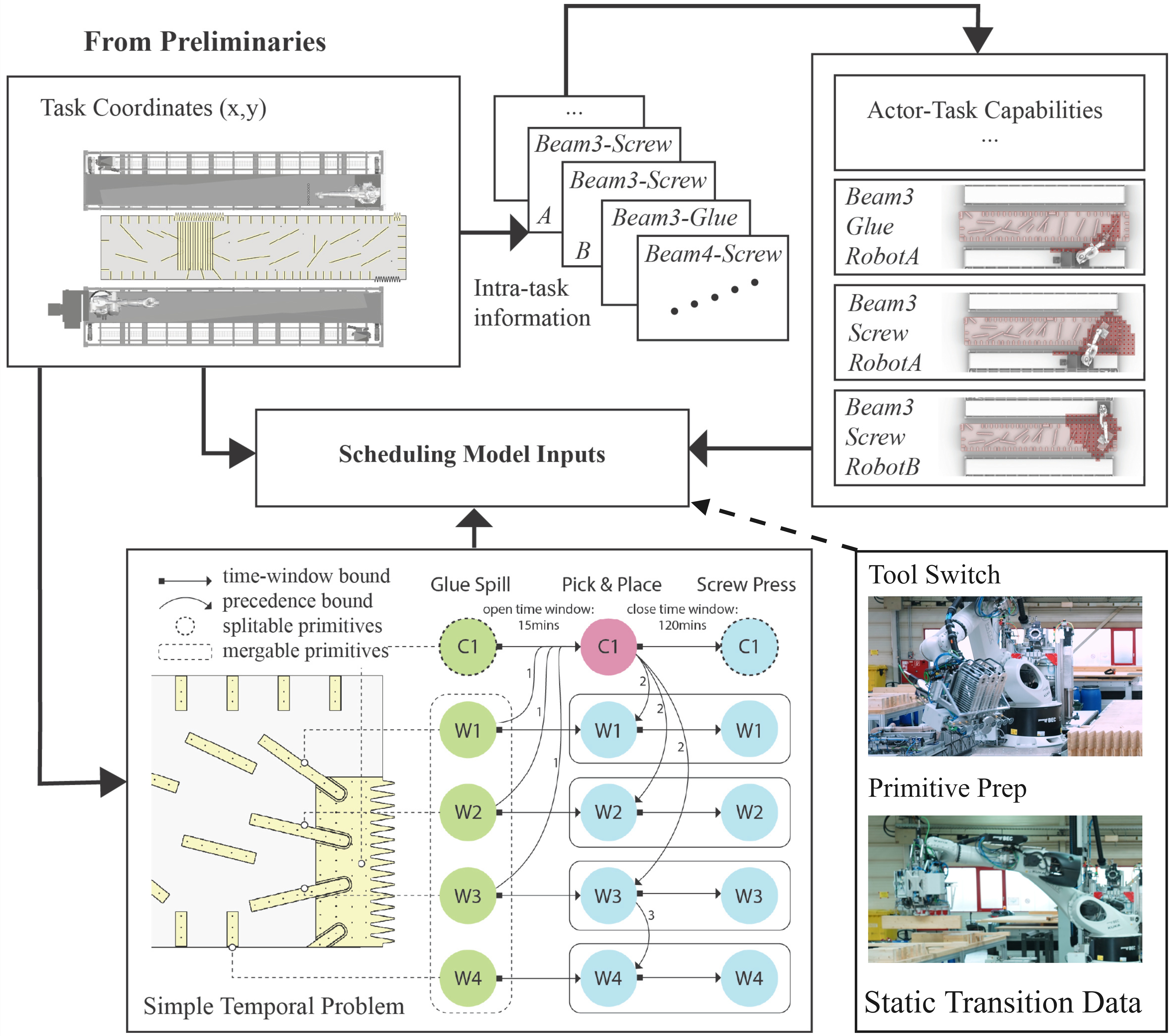}

\caption{Scheduling input summary.}
\label{fig:input}
\end{figure}

\begin{itemize}
\item \textbf{Simple Temporal Problem (STP):} We model the temporal relationships between all tasks as a set of STPs, which define temporal constraints (lower and upper bounds) for task primitives. For each element $e$, the STP enforces that gluing must occur after placement but within the adhesive's open time, and that screwing must occur after placement but within the close time window. For elements $i$ and $j$ with a precedence requirement, a constraint is added to ensure that the placement of $i$ precedes the placement of $j$. 

\item \textbf{Actor-Task Capabilities:} Each task is associated with a specific set of capabilities. This includes the estimated duration for each capable actor to complete the task, robot kinematic states at each task, and a binary representation of its spatial occupancy, crucial for collision avoidance.

\item \textbf{Task Coordinate (x,y):} Each task primitive has one or two (only with gluing task) 2D coordinates in the shared workspace. This information is used to evaluate spatial proximity during the homogeneous screw scheduling, as discussed in Section \ref{sec:prelims}.

\item \textbf{Transition and Tool Switch Information:} The model accounts for logistical overhead, including the time required for tool-switching and preparatory actions like screw loading and material picking. To simplify the spatial problem, these transition tasks are designed to occupy non-colliding areas. The total transition time ($\tau_k^{ij}$) between two consecutive tasks is pre-calculated based on the chosen actor and the sequence of actions. For instance, the transition from a vacuum gripper placement task to a gluing task may include the time for storing the gripper, picking up the glue effector, and pressurizing the adhesive.
\end{itemize}

The problem is defined by the following inputs, variables, and constraints. Let $T$ denote the set of task primitives and $K$ the set of actors. We define a comprehensive set of temporal constraints $\mathcal{T}$ that encapsulates both process precedence and adhesive time windows. Central to our formulation is the concept of a \textbf{level}, formally defined as a set of tasks spatiotemporally disjoint from other levels and synchronized by a global barrier. The \textbf{decision variables} include: the boolean assignments $A_k^i$ for task $i$ to actor $k$; the boolean sequence variables $J_k^{ij}$ (1 if task $j$ immediately follows $i$ on actor $k$); and the integer level index $\ell^i \in \mathbb{Z}_{\ge 0}$ for task $i$. The start/end times $t_S^i, t_E^i$ are modeled as auxiliary variables whose values are propagated based on the main integer decisions. Parameters include the duration $d_k^i$ and transition times $\rho_k^{ij}$. The model minimizes the makespan $C_{\max}$ and the maximum level index $L_{\max}$, weighted by a user-defined factor $\lambda$, as follows:

\begin{gather}
\min\; C_{\max} + \lambda\, L_{\max} \label{eq:obj-clean}, s.t.
\\
\sum_{k\in K} A_k^i = 1 \quad \forall i\in T \label{eq:assign-clean}
\\
A_k^i = 1 \;\Rightarrow\; 
 t_E^i \ge t_S^i + d_k^i
\quad \forall i\in T,\ \forall k\in K
\label{eq:proc-clean}
\\
J_k^{ij}=1 \;\Rightarrow\; t_S^j \ge t_E^i + \rho_k^{ij}
\quad \forall i,j\in T,\ \forall k\in K
\label{eq:routegap-clean}
\\
J_k^{ij}=1 \;\Rightarrow\; \ell^j \ge \ell^i
\quad \forall i,j\in T,\ \forall k\in K
\label{eq:levelmon-clean}
\\
L^{uv} \le t_{\eta_v}^v - t_{\eta_u}^u \le U^{uv}
\quad \forall (u,v)\in \mathcal{T}
\label{eq:stp-clean}
\\
\big(A_a^i \wedge A_b^j \wedge R^{ij}_{ab}=1\big) \;\Rightarrow\; \ell^i \neq \ell^j
\quad \forall i,j\in T,\ \forall a,b\in K
\label{eq:occup-conflict-clean}
\\
\big(J_k^{ij}=1 \wedge R^{(ij)}_{\mathrm{edge}}=1\big) \;\Rightarrow\; \ell^i \neq \ell^j
\quad \forall i,j\in T,\ \forall k\in K
\label{eq:edge-conflict-clean}
\\
t_E^i \le B_{\ell^i}, \qquad B_{k} \le B_{k+1}
\label{eq:barrier-def-clean}
\\
\ell^i < \ell^j \;\Rightarrow\; B_{\ell^i} \le t_S^j
\label{eq:barrier-sep-clean}
\\
C_{\max} \ge t_E^i,\quad L_{\max} \ge \ell^i,\qquad \forall i\in T.
\label{eq:cvars-clean}
\end{gather}

Constraint~\eqref{eq:assign-clean} and \eqref{eq:proc-clean} encode assignment and processing duration. \eqref{eq:routegap-clean}--\eqref{eq:levelmon-clean} enforce routing gaps and level monotonicity along the sequence $J$. \eqref{eq:stp-clean} governs all temporal constraints from $\mathcal{T}$, including precedence (where $L=0, U=\infty$) and adhesive windows via bounded differences. Binary node ($R^{ij}_{ab}$) and edge ($R^{(ij)}_{\mathrm{edge}}$) conflicts are precomputed via voxelized occupancy with safety margins, they parametrize \eqref{eq:occup-conflict-clean}--\eqref{eq:edge-conflict-clean} to lift spatial conflicts into level separation.\eqref{eq:barrier-def-clean}--\eqref{eq:barrier-sep-clean} model timestamp synchronization via barrier variables $B_\ell$. Since these barrier constraints directly bound task start times, Eq.~\eqref{eq:stp-clean} validates temporal window constraints inclusive of synchronization idles. To robustly accommodate execution uncertainty, we apply a safety buffer to the temporal bounds in Eq.~\eqref{eq:stp-clean} by lowering the upper bound $U$, ensuring the schedule remains valid even under stochastic delays. \eqref{eq:cvars-clean} defines the objectives.

%Constraint~\eqref{eq:assign-clean}--\eqref{eq:proc-rec} encode assignment and local processing/preparation. Constraint~\eqref{eq:routegap-clean}--\eqref{eq:levelmon-clean} route gaps and level monotonicity along actor sequences. \eqref{eq:prec-clean} - ~\eqref{eq:stp-clean} process precedence and adhesive windows via bounded differences. Binary node ($R^{ij}_{ab}$) and edge ($R^{(ij)}_{\mathrm{edge}}$) conflicts are precomputed via voxelized occupancy and swept-volume overlap with fixed safety margins; they parametrize \eqref{eq:occup-conflict-clean}--\eqref{eq:edge-conflict-clean} and lift node/edge spatial conflicts into level separation. \eqref{eq:occup-bounds-clean} ties occupancy to task times.  \eqref{eq:barrier-def-clean}--\eqref{eq:barrier-sep-clean} define barrier ordering and objective aggregates. Note that coupled tasks (e.g., glue and screw) are not strictly forced to reside in the same level with the temporal constraints in Eq. \eqref{eq:stp-clean} functioning together with the barrier ordering to encourage grouping them in near levels such that the barrier wait time does not violate the adhesive window. Finally, \eqref{eq:cvars-clean} defines the makespan and the highest level used.

\section{Scheduling Algorithms for Screw-Press Gluing Timber Slab Assembly}
\label{sec:algo}
The formulation presented in the previous section specifies the feasibility and optimality criteria for our problem. However, a monolithic CP model faces significant scalability challenges, particularly for a complete slab assembly that may involve over 800 task primitives. Our key insight is that the timber slab assembly process consists of two fundamentally different scheduling problems.

The \textbf{bottom session} is a \textit{sparse, heterogeneous, constraint-driven} problem. The number of decision nodes is moderate (under 80 in all test cases), but they are linked by complex temporal and logical constraints (e.g., glue windows, tool changes). Conversely, the \textbf{top session} is a \textit{dense, homogeneous, routing-driven} problem, with hundreds of identical screw tasks where path efficiency and workload balance are dominant.

\begin{figure}[h!]
\centering
\includegraphics[width=1.0\linewidth]{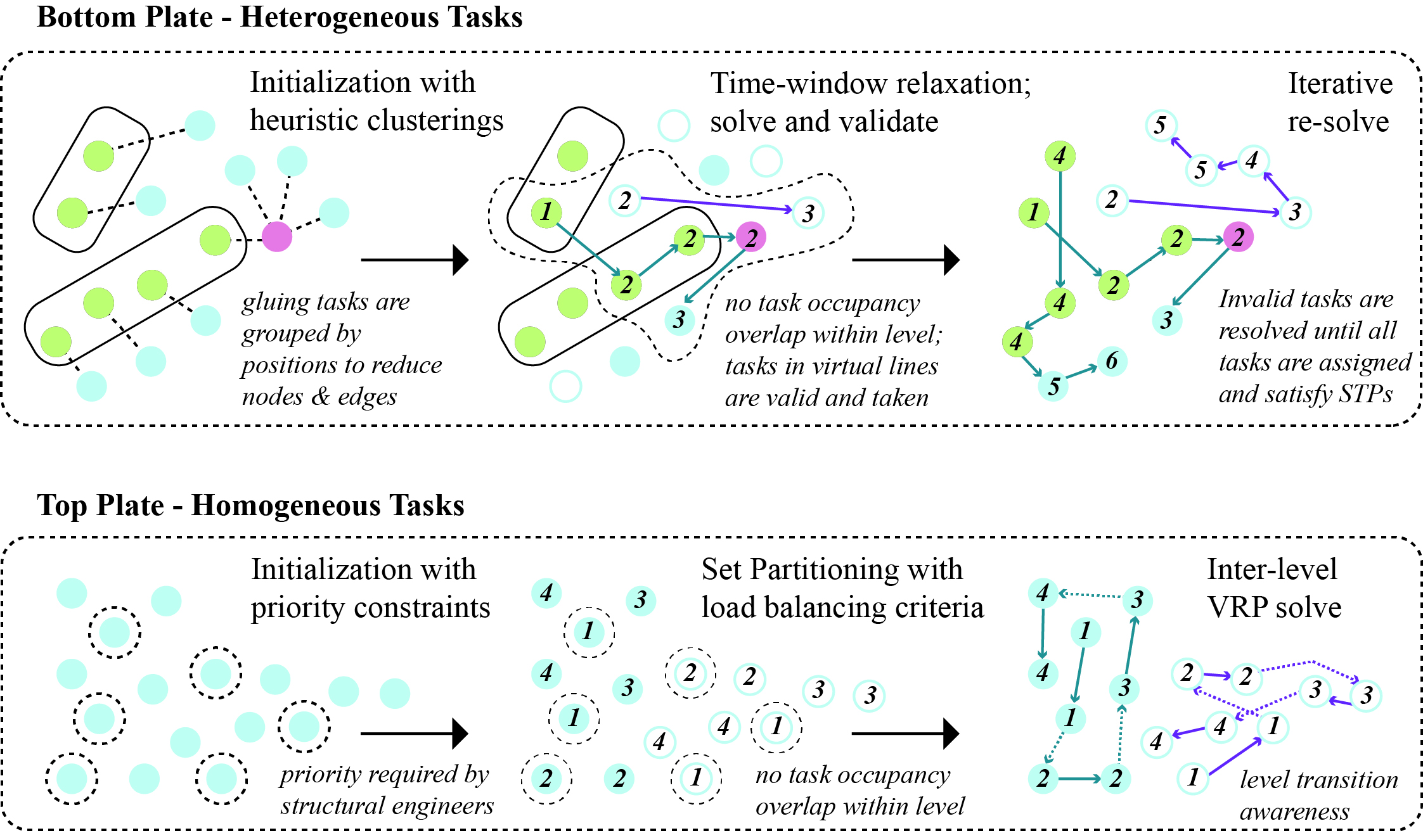}

\caption{Diagram illustration of our algorithms. Bottom Plate: Heterogeneous scheduling with clustering and iterative temporal relaxation. Top Plate: Homogeneous scheduling with decomposed Set Partitioning and Vehicle Routing Problem (VRP) solvers.}
\label{fig:algo}
\end{figure}

The computational study on Fig. \ref{fig:triptych}(a) confirms that a direct CP model of this problem becomes intractable beyond ~80-100 tasks. We therefore dictate a hybrid strategy (Fig.~\ref{fig:algo}): for the constraint-rich bottom session that falls within this tractable range, we employ a direct CP method with primitive clustering and iterative temporal relaxation. For the dense top session, a decomposition approach is necessary and operable under a fixed number of levels.

\subsection{Bottom session: Primitive Clustering and Iterative Temporal Relaxation}
The assembly of the bottom session involves a large number of discrete timber elements that must be glued, placed, and pressed onto a large CLT plate. The ideal sequence of operations to minimize tool changes would involve performing all gluing tasks first, followed by all placement tasks for large elements (using a vacuum gripper), and finally all screwing tasks for smaller linear elements (using a screw effector). Even the tight open-time window of the adhesive (e.g., 15 minutes at $20^\circ$C) makes a single round of such a sequence impossible, we can leverage the inherent efficiency of clustering tasks while systematically addressing the strict temporal windows as shown in Algorithm \ref{alg:bottom_session}:

\begin{algorithm}[h!]
\caption{Iterative Clustering and Scheduling for Bottom Session}
\label{alg:bottom_session}
\begin{algorithmic}[1]
\STATE \textbf{Input:} Task Primitives $T_{all}$, Actor Set $K$, Temporal Constraints $\mathcal{T}$, Collision $R$
\STATE \textbf{Output:} Schedule $S_{final}$

\STATE $(T_{active}, T_{deferred}) \gets \text{InitializeBatches}(T_{all})$

\STATE $\mathcal{T}_{relaxed} \gets \text{ApplyRelaxation}(\mathcal{T})$
\STATE $V \gets \{ \text{dummy} \}$

\WHILE{$V \neq \emptyset$}
    \STATE $S_{skeleton} \gets \text{SolveCP}(T_{active}, \mathcal{T}_{relaxed}, K, R)$
    \STATE $V \gets \text{FindViolations}(S_{skeleton}, \mathcal{T})$
    
    \IF{$V \neq \emptyset$}
        \STATE $T_{active} \gets \text{SplitBatches}(T_{active}, V)$
    \ENDIF
\ENDWHILE

\STATE $S_{final} \gets \text{InsertDeferredTasks}(S_{skeleton}, T_{deferred}, R)$
\STATE \textbf{Return} $S_{final}$

\end{algorithmic}
\end{algorithm}

We first coalesce consecutive glue primitives into clustered batched nodes, reducing the number of decision variables and minimizing travel overhead.  During this aggregation, we relax the temporal bounds associated with the batched glue tasks to permit a feasible initial route.  Three critical screw primitives (the center and end screws) are merged into each placement task to ensure immediate pressing and prevent slippage from the glue line.  All remaining screws are temporarily deferred.  The relaxed instance is solved; any elements whose glue--place chains violate the strict temporal windows are identified, and their glue primitives are detached to form smaller batches.  This process repeats on the unresolved subproblem until every element satisfies its exact (unrelaxed) bounds.  Finally, we reinsert the deferred screws under the full set of assignment, routing, and level constraints.  These deferred tasks are placed into idle slots between levels with fixed actor assignments, maintaining spatial safety.

\subsection{Top session: Set Partitioning and Vehicle Routing Problem}
The top session assembly involves applying glue to the sandwiched timber elements before a top CLT plate is placed via a crane, followed by a dense screw pattern to ensure a secure bond. The multi-robot collaboration must be highly efficient to complete all tasks before the adhesive's strict close time window (approximately 120 minutes under standard conditions).

To address this challenge, we leverage the homogeneous nature of the screwing tasks and reframe the problem as an assignment problem aimed at workload balance. Instead of minimizing makespan as the primary objective in \eqref{eq:obj-clean}, we adopt a load-balancing objective inspired by the \textbf{Multiway Number Partitioning Problem}. This approach seeks to minimize the workload difference between the actors within each level while minimizing the total number of levels to maximize parallelism. For a system with $N$ actors, the general objective is defined as:
\begin{gather}
{g} = \max_{i \in K} \sum_{j \in T_i} w_j - \min_{i \in K} \sum_{j \in T_i} w_j \label{eq:partition_obj}
\end{gather}
where $K$ is the set of actors, $T_i$ is the set of tasks assigned to actor $i$, and $w_j$ is the processing time of task $j$. This equilibrium is achievable for screwing tasks because each primitive's duration is nearly invariant for a given actor, regardless of its sequence, as screw loading occurs in parallel with travel. For our setup in the top session, the KUKA KR210's screwing cycle is 11s (6s screwing and 5s loading), and the KUKA KR420's is 16s (11s screwing and 5s loading), indicating a distribution of 16:11 leads to temporal optimality. We allow a small time difference threshold, $\delta$:
\begin{gather}
{g} \le \delta \label{partition_equilibrium}
\end{gather}
as a constraint for more search space for the solver.

We solve this problem using a two-stage approach, as detailed in Algorithm \ref{alg:top_session}:
\begin{enumerate}
\item \textbf{Task Assignment (Multiway Number Partitioning Problem):} We begin by attempting to assign all tasks with a minimal number of levels, $L_{\max} = k_{min}$, where $k_{min} \ge 2$ to enable multi-robot cooperation. A CP model is built that satisfies the processing, sequencing, level-monotonicity, barrier, and STP constraints described in Sec.~\ref{sec:problem}. To enlarge the search space, we temporarily relax the edge-conflict constraints while retaining the static occupancy constraints. We also introduce a heuristic to prevent isolated assignments: each task assigned to a level must have at least one other task on that level within a distance threshold $\delta_{\mathrm{dist}}$.  This is enforced via additional binary variables linking proximity to assignment; we omit the detailed encoding for brevity.  We relax constraint \eqref{eq:assign-clean} to allow for partial assignment and maximize the number of assigned tasks. Starting with the priority screw set, we solve for $k_{min}=2$. Any unassigned tasks are added to the reinforcement set and solved together with $k_{min}=2$. We find in Sec. \ref{sec:results} that this fixed 2-level solve yields at least 95\% task assignment within a modest time budget.

    \item \textbf{Routing Refinement (Vehicle Routing Problem):} With tasks assigned to specific actors and levels, we solve a multi-vehicle routing problem on each level to ensure collision-free travel. A local search algorithm is implemented on each level, where the start points of each actor's path are fixed by the previous level's solution. This makes each sub-problem a series of fast and scalable vehicle routing problem with fixed candidates. The objective is to minimize the total travel distance, while satisfying the edge-based spatial conflict constraint \eqref{eq:edge-conflict-clean} that was relaxed during the assignment phase. The unassigned screws from the first stage are integrated post-routing by relaxing the workload equilibrium constraint \eqref{partition_equilibrium} and dynamically inserted into existing scheduled levels or added to a minimal number of new levels (e.g., 1--2) to minimize the final makespan.
\end{enumerate}
\begin{algorithm}[h!]
\caption{Bi-Level Scheduling for Top Session Screwing}
\label{alg:top_session}
\begin{algorithmic}[1]
\STATE \textbf{Input:} Task sets $[T_{\text{prio}}, T_{\text{reinf}}]$, Actors $K$, Collision data $R$, Load-balance threshold $\delta$
\STATE \textbf{Output:} Completed Schedule $S_{\text{final}}$ (Task Assignments, Orders, Levels)
\STATE Initialize $S_{\text{assignments}} \gets \emptyset$, $T_{\text{unassigned}} \gets \emptyset$, $L_{\max} \gets 2$
\FOR{each $T_{\text{set}}$ in $[T_{\text{prio}}, T_{\text{reinf}}]$}
    \STATE $T_{\text{current\_solve}} \gets T_{\text{set}} \cup T_{\text{unassigned}}$
    \STATE $S_{\text{partial}} \gets \text{SetPartition}(T_{\text{current\_solve}}, L_{\max}, \delta)$
    \STATE $T_{\text{newly\_assigned}} \gets \text{GetAssignedTasks}(S_{\text{partial}})$
    \STATE $T_{\text{unassigned}} \gets T_{\text{current\_solve}} \setminus T_{\text{newly\_assigned}}$
    \STATE $S_{\text{assignments}} \gets S_{\text{assignments}} \cup S_{\text{partial}}$
\ENDFOR
\STATE $S_{\text{initial\_routes}} \gets \text{Initialize empty schedule}$
\STATE $p_{0} \gets \text{-1}$ \COMMENT{Start index at level 0 is not forced}
\FOR{each level $l$ from $0$ to $L_{\max}-1$}
    \STATE $T_l  \gets \text{Tasks in } S_{\text{assignments}} \text{ at level } l$
    \STATE $T_{l+1}  \gets \text{Tasks in } S_{\text{assignments}} \text{ at level } l+1$
    \STATE $\text{S}_l, p_{l+1} \gets \text{SolveVRP}(T_l, T_{l+1}, p_l)$
    \STATE Add $\text{S}_l$ to $S_{\text{initial\_routes}}$
\ENDFOR
\STATE $S_{\text{final}} \gets \text{InsertUnassigned}(S_{\text{initial\_routes}}, T_{\text{unassigned}})$ 
\STATE \textbf{Return} $S_{\text{final}}$
\end{algorithmic}
\end{algorithm}

\section{Level-Based Task Execution}
\label{sec:execution}

We employ a semi-centralized control framework, \textit{Twico} \cite{Skoury2024DigitalTA}, to orchestrate the system. While general-purpose middlewares like MoveIt 2 \cite{MoveIt2Async2024} offer asynchronous capabilities, \textit{Twico} is specifically tailored to dispatch scheduled primitives as parameterized KUKA subprogram calls with level-based barrier protocol for spatial safety, ensuring seamless alignment with our task scheduling model.

Before execution, the scheduled primitives are post-processed into executable task objects. To optimize efficiency, consecutive tasks of the same type (e.g., multiple screwing points or continuous glue lines) are batched into single subprogram calls, and global task poses are transformed into each robot's local coordinate frame. These objects, defined within a unified task schema \cite{SKOURY2024105229} including task ID, level index, and subprogram parameters, are stored on the central server and dispatched to parallel execution threads.

In the physical layer, each robot's controller interprets the task parameters into KRL subroutines and executes the required motions and skills. The controller itself monitors primitive completion through internal signal-based commands and reports status back to the server. The server functions as a global barrier manager: it tracks the completion status of the current level and only releases tasks for level $\ell+1$ once all actors have confirmed the completion of level $\ell$. This protocol acts as a robust control law that absorbs timing uncertainty and ensures collision-free operation without the need for dense timestamp synchronization.

\section{Framework Validation and Performance}
\label{sec:results}

We validate our framework through three studies: first, by successfully fabricating a full-scale, complex architectural prototype; second, by highlighting the robustness embedded in our framework's design; and third, by demonstrating the scalability and necessity of our tailored hybrid algorithm on a representative set of design variants.

\subsection{Physical Prototype Fabrication}

The 2.4\,m$\times$6\,m slab (34 elements; 352 screws) executed collision-free under adhesive windows using 15 levels (11 bottom, 4 top) and 108 subroutines. Realized makespans were 28/20\,min vs.\ model estimates 25/19\,min (both within 5\%). The physical execution, while successful, highlighted the importance of our level-based barrier mechanism. Minor variations in material handling time and screw-driving duration (up to 5\% deviation from the model's estimates) were naturally absorbed within each level's asynchronous execution window, without requiring re-planning or risking collisions. This real-world validation confirms the framework's inherent robustness against the uncertainties of physical fabrication. 

\begin{figure}[h!]
    \centering
     \includegraphics[width=1.0\columnwidth]{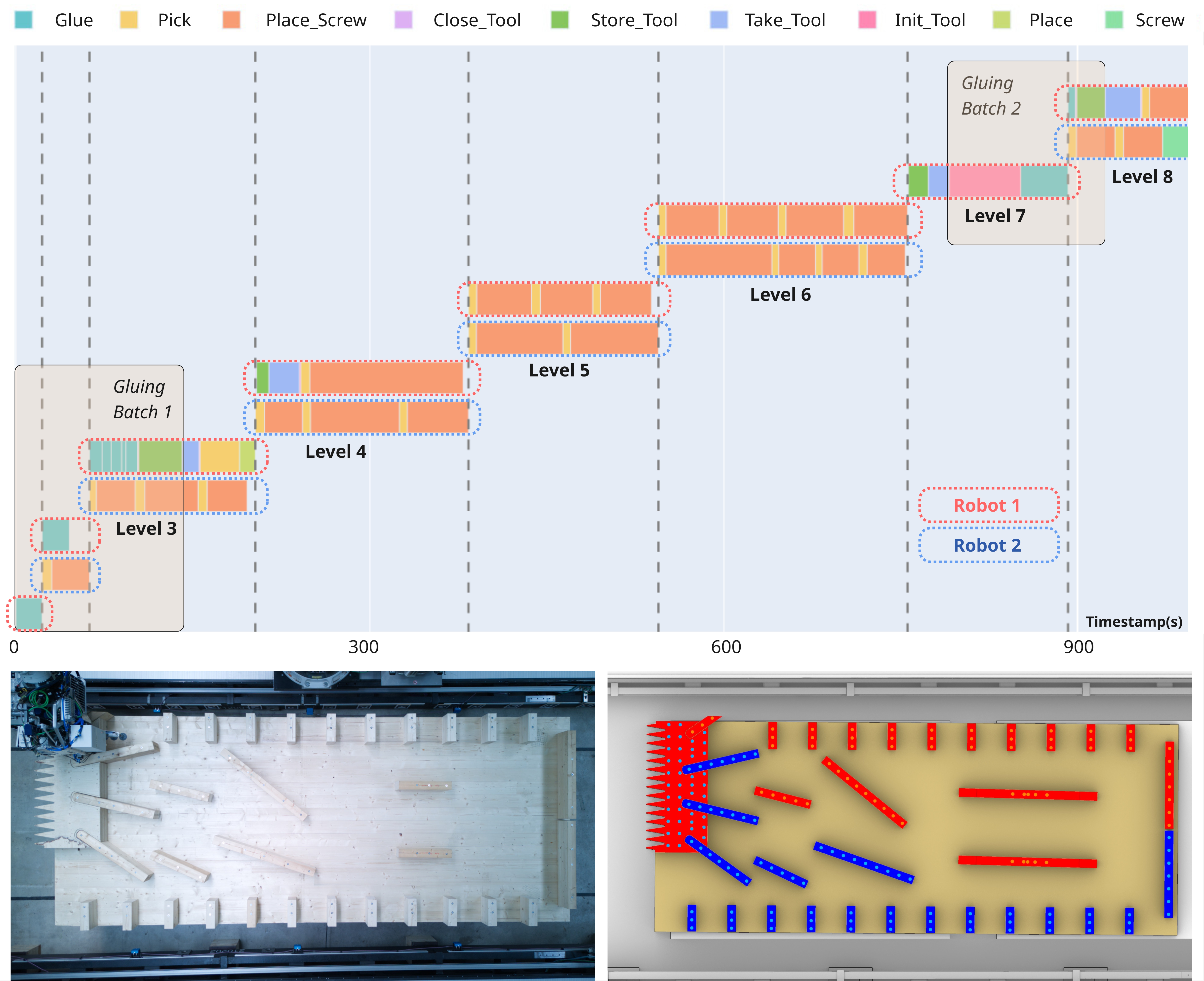}
    \caption{Overview of bottom session of our prototype. Top: Execution Gantt Chart from the first 990s up until level 8. Bottom left: Picture of the full assembly. Bottom right: Task assignment diagram to the robots.}
    \label{fig:bottom_overview}
\end{figure}

\begin{figure}[h!]
    \centering
     \includegraphics[width=1.0\columnwidth]{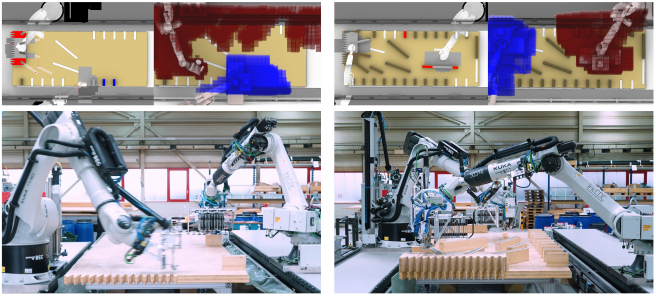}
    \caption{Analysis of Level 2 (left) and Level 8 (right) of the bottom session. Specifically, Level 2 consists gluing, tool switching, and pick-and-place of a crown element for Robot 1, while Robot 2 is able to execute pick-and-place and screwing tasks for 3 smaller beams, leading to efficient task distribution.}
    \label{fig:bottom_level}
\end{figure}

\begin{figure}[h!]
    \centering
     \includegraphics[width=1.0\columnwidth]{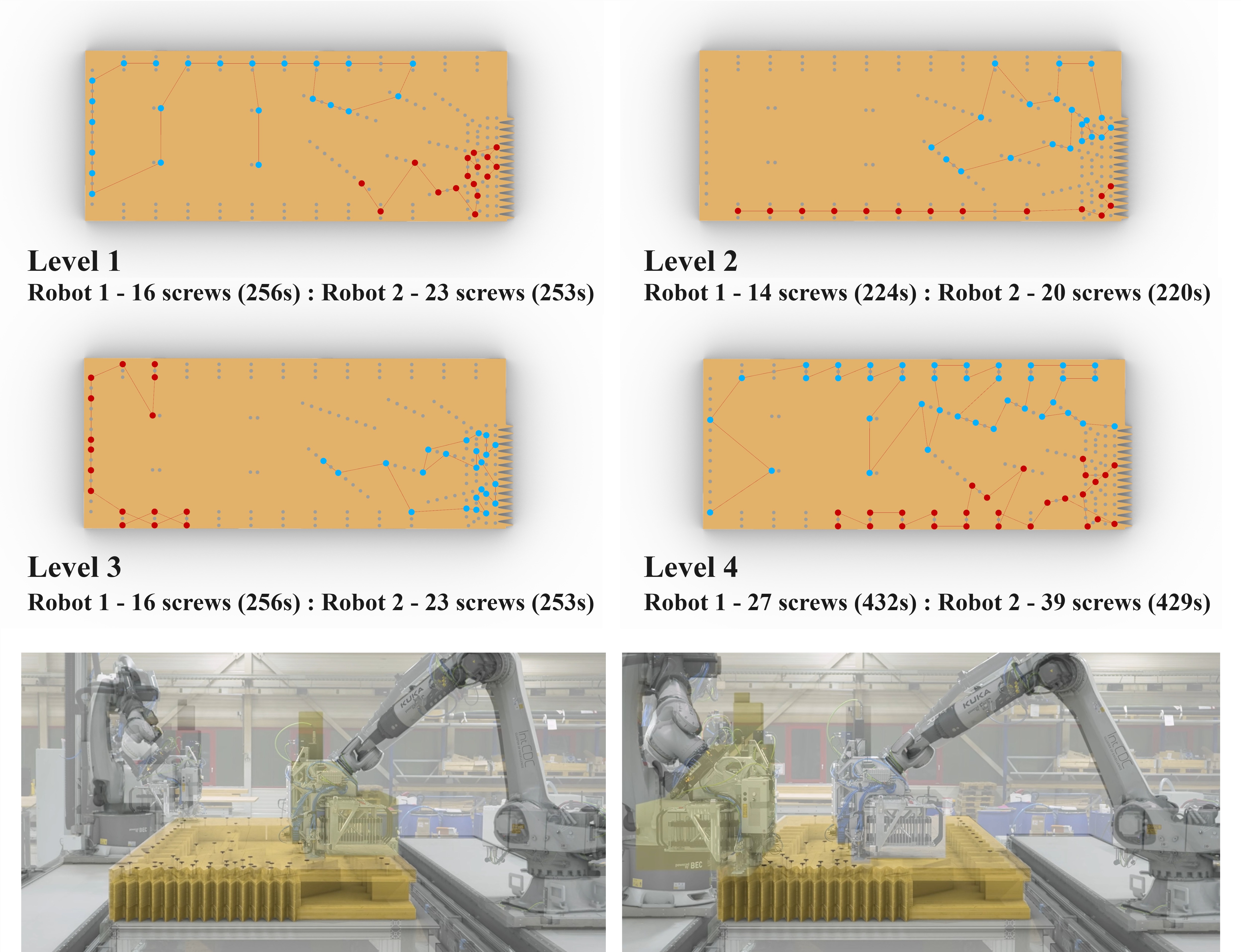}
    \caption{Scheduling result (top) and execution process with simulation overlay (bottom) of the top session. A total of 4 levels are scheduled with an estimated 98.8\% robot utilization ratio.}
    \label{fig:top_result}
\end{figure}

Fig. ~\ref{fig:bottom_overview} summarizes the bottom assembly session involving heterogeneous task cooperation, such as gluing application, placement of large timber crowns and small timber beams, as well as screwing. Fig. ~\ref{fig:bottom_level} provides a detailed analysis on two scheduled levels from the bottom session. Fig. \ref{fig:top_result} demonstrates the scheduling and execution of the top screwing session, which contains 4 levels for 178 screws while maintaining workload balance within each level.

\subsection{Robustness to Execution Variability}
We tested schedule robustness by injecting zero-mean i.i.d.\ noise into per-primitive durations and tool-switch times (Gaussian and bounded uniform), re-simulating the controller–dispatcher loop in a virtual environment provided by KUKA.OfficeLite. Across noise levels up to \textit{5\%}, all runs respected level barriers and adhesive windows. Beyond simulated robustness, our framework incorporates two key mechanisms at the design level to handle real-world uncertainties:

\paragraph{Proactive Safety Buffers} The scheduler operates on conservative time windows. For instance, a hard 15-minute adhesive open-time is translated into a softer 13.5-minute scheduling constraint (a 10\% safety buffer), ensuring minor delays do not compromise process integrity.

\paragraph{Inherent Fault Tolerance} The level-based barrier protocol provides a powerful fault-tolerance mechanism. A task failure or hardware malfunction on one robot simply pauses the system at a safe, collision-free state upon completion of the current level. This creates a natural checkpoint for human intervention without jeopardizing the other robot, a critical feature for long-horizon autonomous construction.

\begin{figure}[t]
  \centering
  \includegraphics[width=\columnwidth]{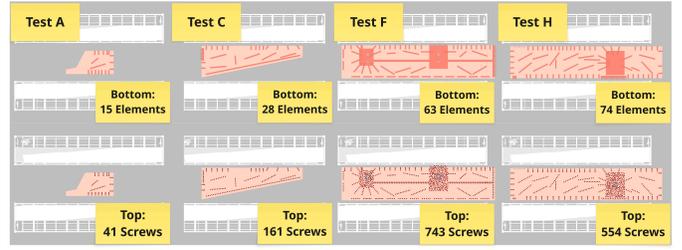}
  \caption{Four of the eight representative slab geometries used for the computational study, illustrating the significant diversity in element count (15 to 74) and screw density (41 to 743). A detailed table of per-test metrics is available in the supplementary material.}
  \label{fig:design_variants}
\end{figure}

\subsection{Scalability and Performance}

To analyze the framework's performance across a range of complexities, we evaluated it on 8 distinct slab geometries sourced from a real-world building project (Fig.~\ref{fig:design_variants}). These cases represent a significant diversity in scale, with 15 to 74 internal elements and 41 to 743 screws.

\begin{figure}[t]
  \centering
  \includegraphics[width=\columnwidth]{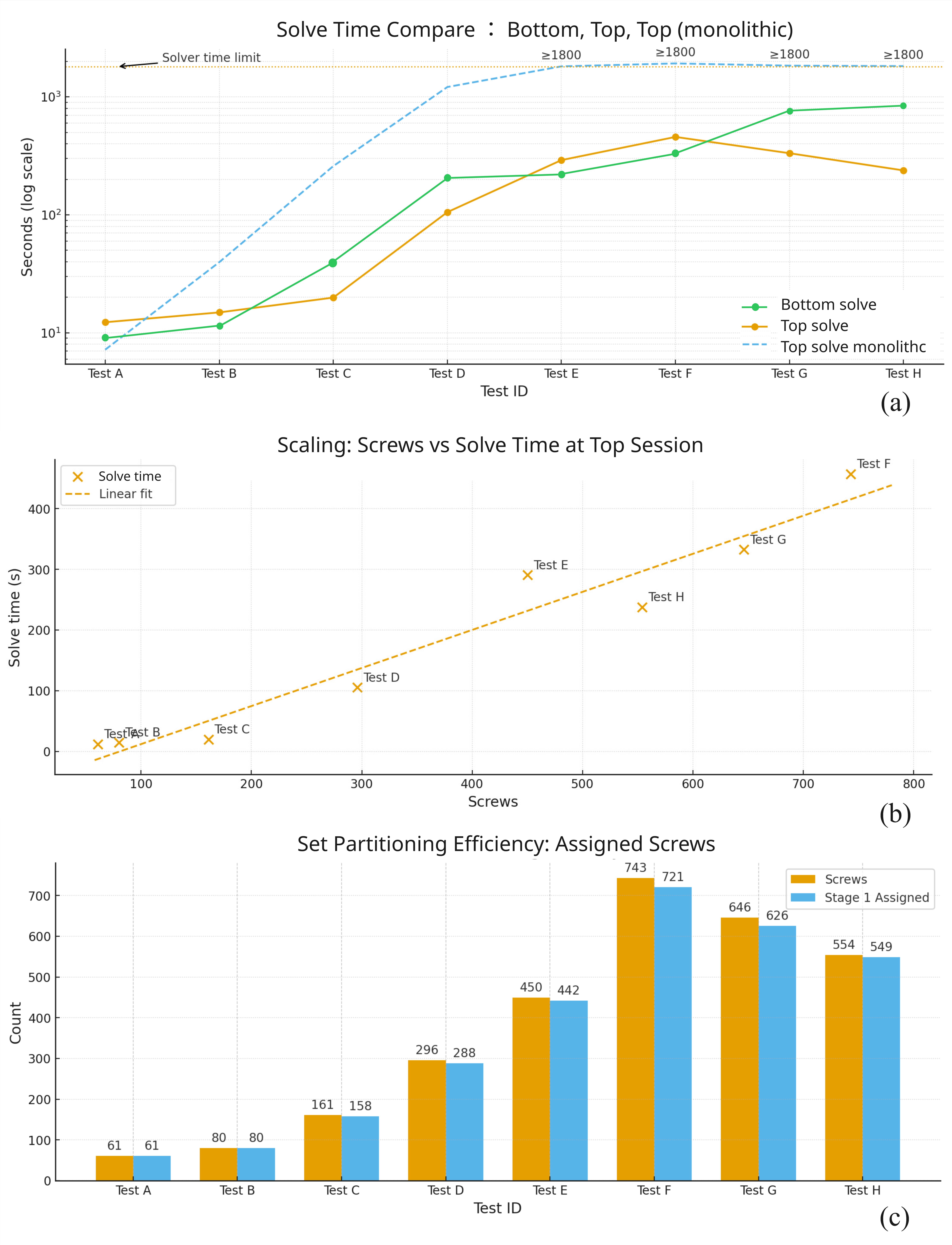}
  \caption{Scaling, ablation and makespan overview: (a) solve time compare: bottom, top and top(monolithic); (b) top-session screws vs.\ solve time; (c) efficiency at set partition stage for top session solve. }
  \label{fig:triptych}
\end{figure}

Our central finding is that a monolithic CP formulation becomes sensitive to problem size, shown in Figure~\ref{fig:triptych}(a). The monolithic approach for the top session problem fails to find a solution within the 1800s time limit for instances with more than ~300 tasks (Slabs E-H). This data-driven insight is the direct justification for our hybrid algorithm strategy for top and bottom sessions.

\paragraph{For the Bottom Session} After our primitive clustering, the number of decision nodes is identical to the element count (max 74 in Test H). This count falls within the tractable range for a direct CP solver. Therefore, our heterogeneous algorithm model assignment, sequence and levels as variables directly in the CP model with only relaxed temporal constraints, as it optimally minimizes the level count under the spatial disjunction of the sub-problem. Results from Figure~\ref{fig:triptych}(a) indicates that the problem is still solvable at its maximum size under the context of the studied cases.

\paragraph{For the Top Session} The number of tasks (screws) is an order of magnitude larger, far exceeding the CP solver's tractable limit. This necessitates the decomposition into a set-partitioning problem for load-balancing and a VRP for routing. As shown in Figure~\ref{fig:triptych}(b), this tailored approach scales effectively, solving even the largest instances within 460 seconds. We also notice from Figure~\ref{fig:triptych}(c) that the initial partitioning model manages to assign at least 98\% of the screws to the load-balanced levels, making the insertion problem in the final step of Algorithm \ref{alg:top_session} easier.

% ---------- Figure: top-session scaling (single column) ----------
% ---------- Figure: ablation (single column; scatter + fits; mono clipped) ----------

\section{Conclusion}
%We have introduced LASER, a level-based scheduling and execution regime designed for spatiotemporally constrained multi-robot manufacturing. We demonstrated its efficacy in the time-critical screw-press gluing process for large-scale timber slab assembly. The resulting schedule allows for asynchronous execution within levels and single synchronization between them. By transforming stringent adhesive windows and spatial safety requirements into a robust, level-based execution structure, our work paves the way for automated robotic fabrication of complex, sustainable timber architecture, with potential extension to broader manufacturing scenarios with dense fastening or curing deadlines. Future work could explore scalable temporal-based decomposition methods for heterogeneous collaboration if a potential assembly design contains larger element counts.

We have introduced LASER, a level-based scheduling and execution framework that enables time-critical process of multi-robot screw-press gluing for large-scale timber slab assembly. The resulting schedule allows for asynchronous execution within levels and single synchronization between them. By transforming stringent adhesive windows and spatial safety requirements into a robust, level-based execution structure, our work paves the way for automated robotic fabrication of complex, sustainable timber architecture. 

By embedding collision avoidance directly into the schedule, our system safely operates without the computational overhead of continuous collision checking. While such a mechanism efficiently coordinates sparse tasks, highly crowded workspaces may reduce the collaboration efficiency or cause levels to degenerate into serialized single-robot tasks and reduce parallelism. Future work could explore scalable temporal-based decomposition methods for heterogeneous collaboration with larger element counts and hybridize level-barriers with temporal networks to reduce idle times.

\bibliographystyle{IEEEtran}
\bibliography{references}

\end{document}